\documentclass[10pt,twocolumn,letterpaper]{article}

\usepackage{cvpr}              

\usepackage[dvipsnames]{xcolor}

\definecolor{cvprblue}{rgb}{0.21,0.49,0.74}
\usepackage[pagebackref,breaklinks,colorlinks,citecolor=cvprblue]{hyperref}

\def\paperID{*****} 
\def\confName{CVPR}
\def\confYear{2024}

\title{Efficient Video-Based ALPR System Using YOLO and Visual Rhythm}

\author{Victor N. Ribeiro and Nina S. T. Hirata\\
Instituto de Matemática e Estatística - Universidade de São Paulo\\
SP - São Paulo, Brasil\\
{\tt\small victor\_nascimento@usp.br, nina@ime.usp.br}
}

\begin{document}
\maketitle
\begin{abstract}
Automatic License Plate Recognition (ALPR) involves extracting vehicle license plate information from image or a video capture. These systems have gained popularity due to the wide availability of low-cost surveillance cameras and advances in Deep Learning. Typically, video-based ALPR systems rely on multiple frames to detect the vehicle and recognize the license plates. Therefore, we propose a system capable of extracting exactly one frame per vehicle and recognizing its license plate characters from this singular image using an Optical Character Recognition (OCR) model. Early experiments show that this methodology is viable.
\end{abstract}    
\section{Introduction}
\label{sec:intro}

Automatic License Plate Recognition (ALPR) is widely used in road traffic monitoring, automatic toll collection and other applications~\cite{laroca, survey}. This task can be difficult due to the varied formats of license plates and inconsistent outdoor lighting conditions in video captures~\cite{alpr-difficulties}. Thus, there is still a demand for research on license plate recognition systems and algorithms to enhance their performance~\cite{alpr-demand}. 

Capturing a clear image of a moving vehicle with a visible license plate poses a significant challenge~\cite{vehicle-video}. In real-world scenarios, video-based ALPR systems rely on multiple frames rather than a single image~\cite{alpr-multiple-frames}.

In this paper, we present a more efficient approach for ALPR in videos. Our approach combines YOLO, a established object detection model, with Visual Rhythm, a technique for generating time-spatial images from videos. By merging these methods, we can selectively extract and process a single frame per vehicle, facilitating license plate character recognition using OCR models based on this singular image. 


\section{Background}
\label{sec-bg}

In this section, we provide essential background information for our proposed ALPR system. 

\subsection{YOLO}

YOLO (You Only Look Once)~\cite{yolo-v1} is a state-of-the-art object detection deep learning model. The latest version, YOLOv9, has improved by integrating superior feature extraction capabilities, leading to more precise detections~\cite{yolo-evo, yolov9}.

\subsection{Visual Rhythm}

Visual Rhythm (VR) efficiently gathers spatial and temporal information, condensing the video content into a single image~\cite{VR-definition}. This technique facilitates the identification of frames featuring crucial visual content, thereby reducing the computational complexity tied to frame-by-frame processing~\cite{vr-video}, at the cost of sacrificing real-time processing.

\begin{figure}[htp]
    \centering
    \includegraphics[width=1\linewidth]{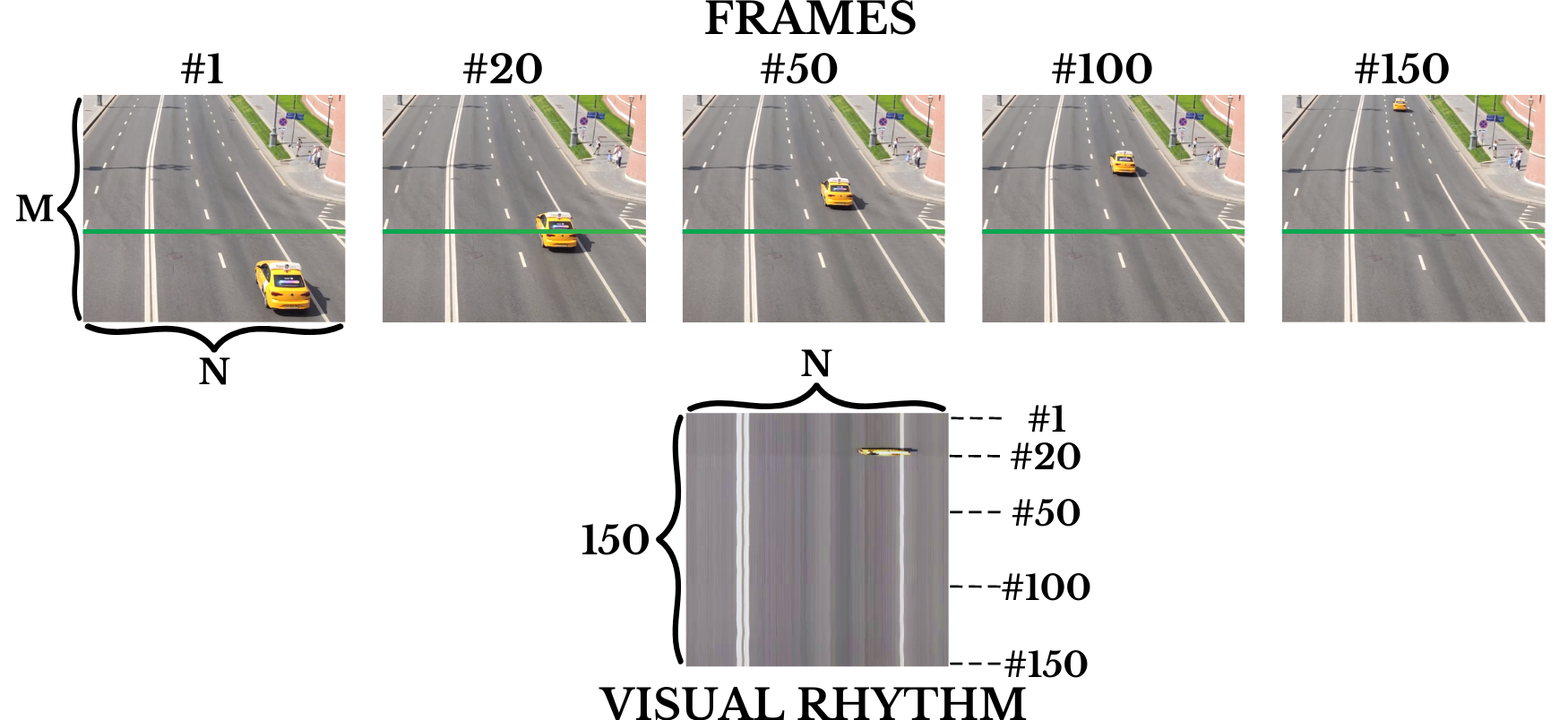}
    \caption{Visual Rhythm Generation.}
    \label{fig:vr-generation}
\end{figure}

Consider a video denoted as $f$ with $T$ frames of size $M \times N$. The VR method is applied to each frame $f_1, \dots, f_T$, capturing exclusively the pixels along a predefined line. The collection of pixels from frame $t$, $t=1,\dots,T$, on this line are stacked along the time axis to form an image of size $T \times N$, a time-spatial representation of the video, as show in Figure \ref{fig:vr-generation}. The top part shows five frames of a video sequence, where the "line" is superimposed in green (best viewed in electronic format), and the bottom part shows the VR image (time dimension in the vertical axis) of the whole video sequence. Whenever an object crosses the line, a mark is observed in the image, at the row corresponding to the frame index. Each object is represented by a single mark.

Applications using VR assume unidirectionally moving objects that cross the line in a top-view video and at a velocity within the camera's frame rate \cite{VR-plankton}.


\subsection{EasyOCR}

To decode license plate content, we employ EasyOCR~\cite{easyocr}. A model based on the CRAFT algorithm~\cite{craft} for character region awareness in the detection task. For recognition, it adopts a CRNN architecture, using ResNet and LSTM networks, and CTC for decoding~\cite{crnn}.
\section{Methodology}

\begin{figure}[htp]
    \centering
    \includegraphics[width=1\linewidth]{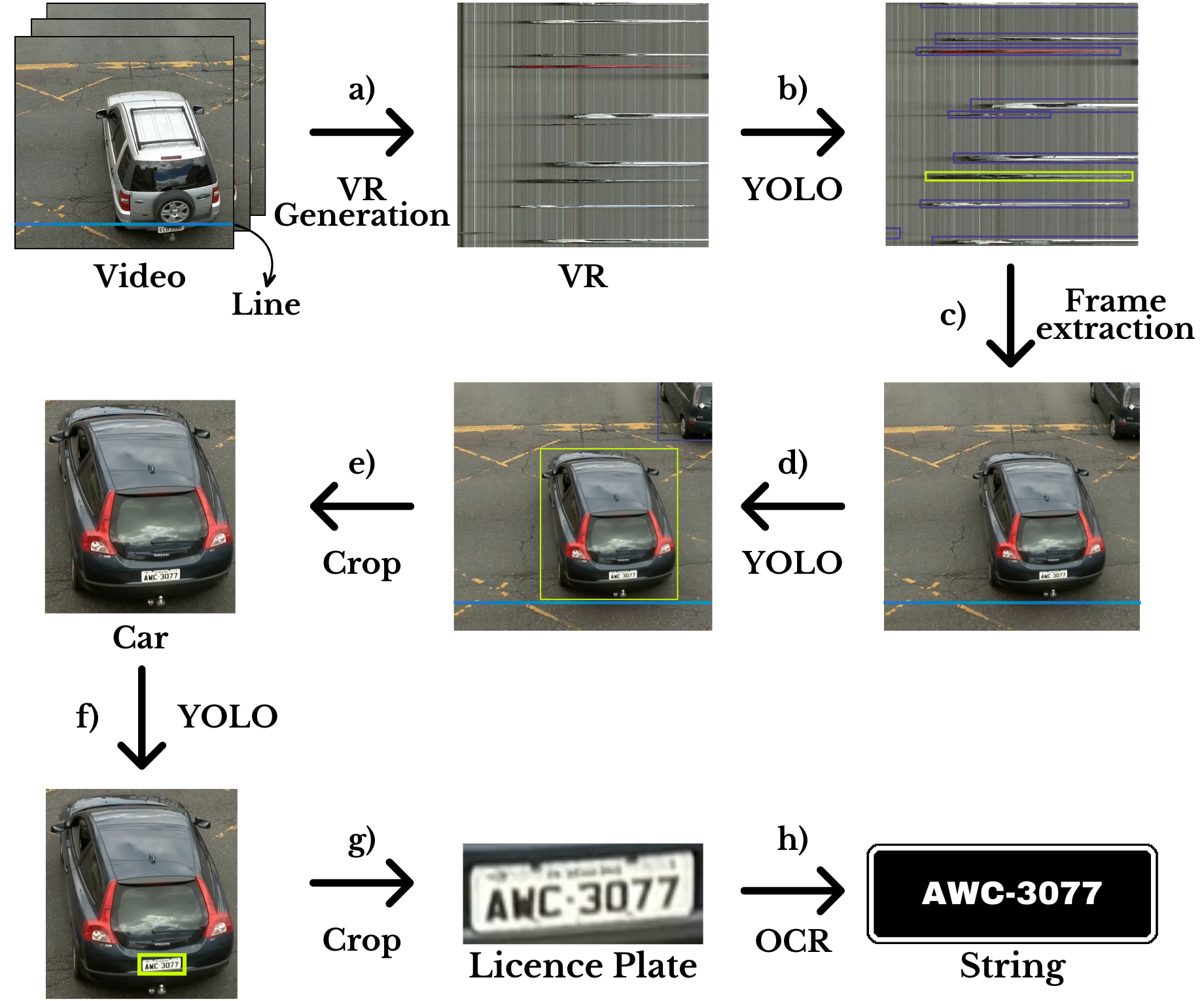}
    \caption{Data flow in the VR–based ALPR.}
    \label{fig:approach}
\end{figure}

The proposed ALPR system is similar to the approach outlined by Ribeiro and Hirata~\cite{sibigrapi}. Our contribution lies in expanding this method to encompass the recognition of vehicle license plates. The main steps of the proposed ALPR system is illustrated in \Cref{fig:approach}.

First, in step (a), we generate a VR image for segments of $T$ consecutive frames (we will use $T = 600$ due to memory limitations). In the next step (b), we employ YOLO to detect each of the marks within the VR image.

Next, step (c), for each detected mark, we extract the corresponding frame from the video. To achieve this, considering each mark as a representation of a vehicle, we can infer that the $y$ coordinate of the mark's bottom corresponds to the temporal index of the frame when the vehicle entirely crosses the line. In~\Cref{fig:approach} the extracted frame corresponds to the mark highlighted in yellow in the VR image.

After obtaining the relevant frame, in step (d), we employ YOLO to detect vehicles in the extracted frame. Following that, in step (e), we identify the specific vehicle represented by the mark by comparing its bounding box $x$ coordinates with the mark coordinates in the VR image.

Continuing, in step (f), YOLO is utilized on the cropped vehicle image to identify the license plate coordinates. Following this, in step (g), we refine the image by cropping it, and in step (h), an OCR algorithm is applied to extract the characters from the license plate. After completing these steps, we repeat for the next $T$ frames until the video ends.
\section{Evaluation and Experiments}

This section explains the experimental setup and provides early experiments of the proposed approach for ALPR.

\subsection{Preparation}

In this study, we utilized YOLOv8-small~\cite{yolov8}, which was pre-trained on COCO and fine-tuned for vehicle and mark detection using a task-specific dataset~\cite{sibigrapi}. For the licence plate detection, we fine-tuned the model in a public available dataset containing more than 10.000 images~\cite{lp-dataset}.

To conduct the experiments for this work, we will use the Vehicle-Rear dataset~\cite{vehicle-rear}, a novel dataset for vehicle identification. This dataset contains a set of videos that meet the constraints required by the VR method, while also featuring vehicle license plates with high resolution. Notably, the license plates in this dataset adhere to the Brazilian format.

For recognizing license plate characters, we employed the pre-trained EasyOCR model.  

\subsection{Early Experiments}

The experiments were conducted exclusively on Video 5 from Camera 2 and were evaluated solely on segments where vehicles move vertically, with the line at $y = 800$. Our approach achieved around $15.76\%$ Character Error Rate (CER) when identifying the license plate characters. 

It's important to note that if one of the early steps fails and becomes impossible to extract the license plate, we consider the approach to have misread every character in the license plate for the CER calculation. 

Given that YOLO was fine-tuned on a separate dataset under different circumstances, and we are using the base EasyOCR model, that supports 80+ languages and hasn't been fine-tuned specifically for recognizing license plate characters, this outcome is relatively good.

\section{Conclusion}

In this paper, we propose a more efficient video-based ALPR system, wherein the license plate characters are identified using a single image per vehicle. Based on preliminary findings, we conclude that the methodology is viable, but there is ample room for improvement. Our next aim is to fully explore this problem to improve computational efficiency and recognition efficacy. Some ideas we have in mind are: training YOLO on the Vehicle-Rear dataset for both mark and vehicle detection, and exploring different OCR models.
{
    \small
    \bibliographystyle{ieeenat_fullname}
    \bibliography{main}
}


\end{document}